\documentclass[10pt,twocolumn,letterpaper]{article}

\usepackage{cvpr}
\usepackage{times}
\usepackage{epsfig}
\usepackage{graphicx}
\usepackage{amsmath}
\usepackage{amssymb}
\usepackage{subcaption}
\usepackage{bm}

\usepackage{enumitem}
\usepackage{booktabs}
\usepackage{algpseudocode,algorithm,algorithmicx}

\usepackage[pagebackref=true,breaklinks=true,letterpaper=true,colorlinks,bookmarks=false]{hyperref}

\cvprfinalcopy 

\def\cvprPaperID{2831} 

\ifcvprfinal\fi
\begin{document}

\title{Evolving Losses for Unsupervised Video Representation Learning}

\author{AJ Piergiovanni, Anelia Angelova, Michael S. Ryoo\\
Research at Google \\
\texttt{\{ajpiergi,anelia,mryoo\}@google.com}\\}

\maketitle

\begin{abstract}
We present a new method to learn video representations from large-scale unlabeled video data. Ideally, this representation will be generic and transferable, directly usable for new tasks such as action recognition and zero or few-shot learning. We formulate unsupervised representation learning as a multi-modal, multi-task learning problem, where the representations are shared across different modalities via distillation. Further, we introduce the concept of loss function evolution by using an evolutionary search algorithm to automatically find optimal combination of loss functions capturing many (self-supervised) tasks and modalities. 
Thirdly, we propose an unsupervised representation evaluation metric using distribution matching to a large unlabeled dataset as a prior constraint, based on  Zipf's law. This unsupervised constraint, which is not guided by any labeling, produces similar results to weakly-supervised, task-specific ones. 
The proposed unsupervised representation learning results in a \emph{single} RGB network and outperforms previous methods. Notably, it is also more effective than several label-based methods (e.g., ImageNet), with the exception of large, fully labeled video datasets.



\end{abstract}

\section{Introduction}

\begin{figure}
    \centering
    \includegraphics[width=\linewidth]{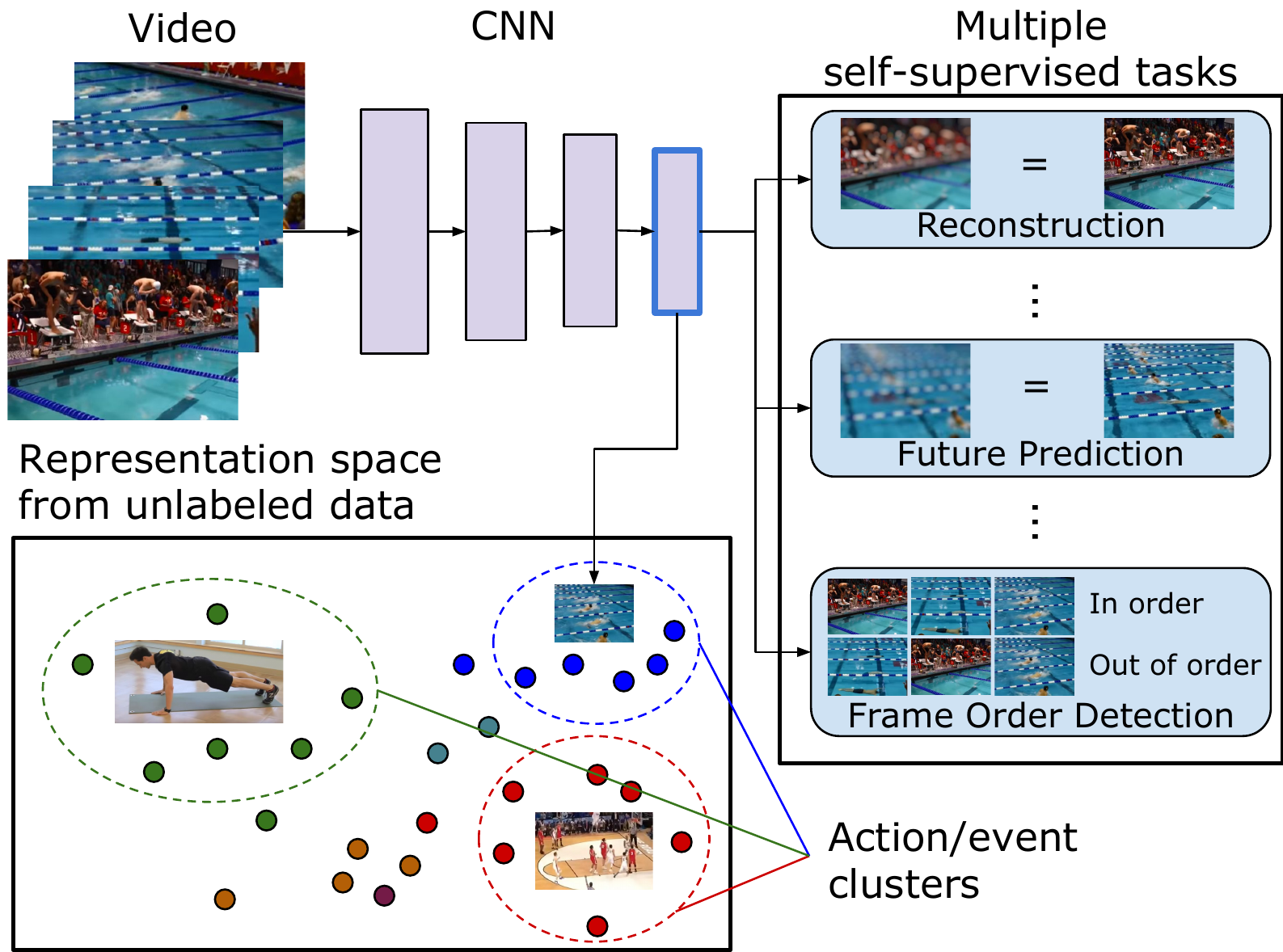}
    \caption{
    Overview of our unsupervised representation learning framework. 
    The objective is to obtain a good representation (blue outlined box) from a set of self-supervised tasks. 
    We use an evolutionary algorithm to automatically find the optimal combination of tasks and power law distribution matching to `supervise' the clustering and guide the evolution. No labeling or supervision is needed.}
    \label{fig:concept}
\end{figure}

Video representation learning is an important problem which benefits high-level perception tasks including action recognition and video object detection~\cite{tran2014c3d,carreira2017quo,tran2018closer}. It has many key applications, such as web-video retrieval, robot perception, and smart homes and cities.
However, learning visual representations generally requires a large number of labeled training examples. This is even more so for videos, as videos are higher-dimensional input than images and video CNNs have more learnable parameters than 2D ones. 
Simultaneously, videos are more expensive to collect and annotate than images, as they require additional, and often ambiguous, temporal annotations \cite{whatactions}. Additionally, in rare event detection, very few examples may be available to begin with.
Thus, obtaining a good video representation without relying on domain specific, annotated video samples, has significant impact on real-world scenarios where large-scale video data curation and labeling is prohibitive.

\begin{figure*}
    \centering
    \includegraphics[width=0.98\linewidth]{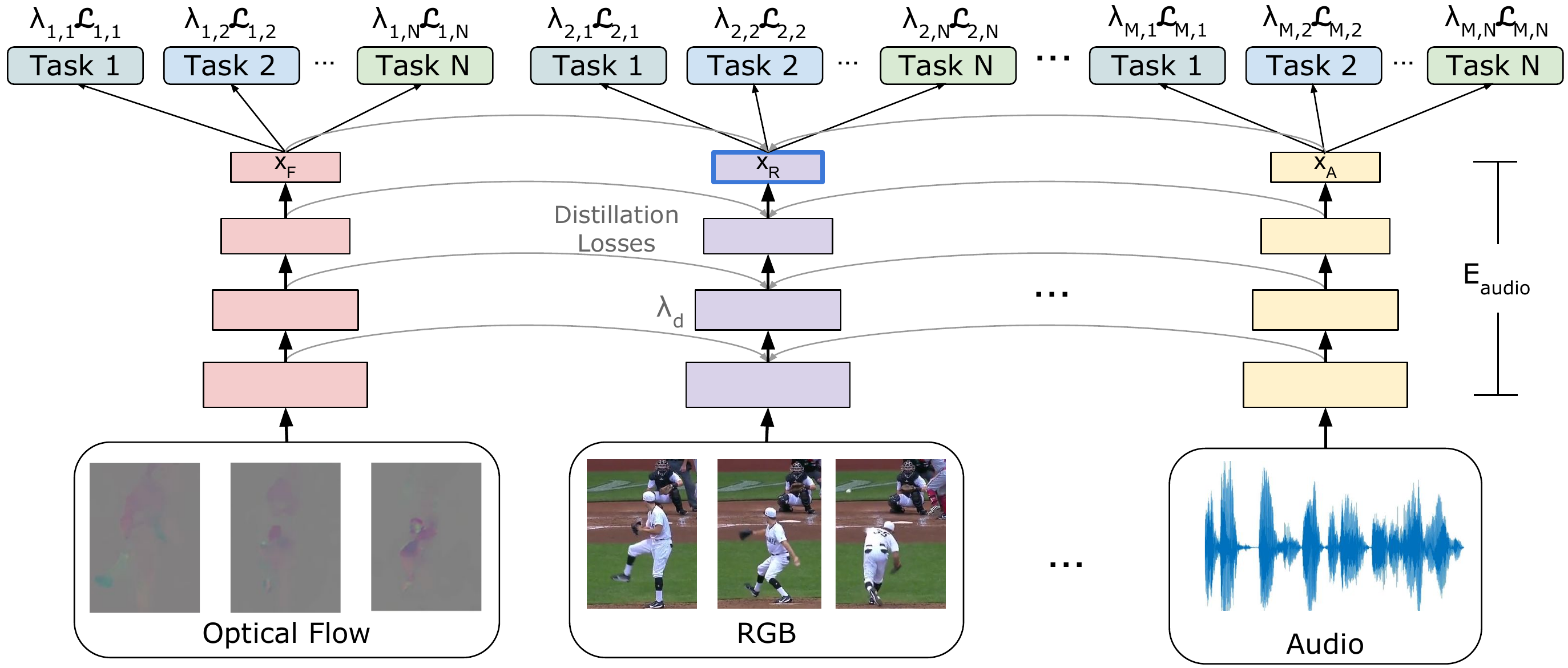}
    \caption{
    The multi-task, multi-modal, unsupervised representation learning framework. Each modality is trained to optimize a set of tasks. 
    Distillation regularization loss terms `infuse' each modality's information into the main RGB network (drawn center). We evolve the loss function to automatically find the weights for each task and distillation location, via an unsupervised objective. The goal is to obtain representation from a \textbf{single} RGB network that transfers to recognition tasks.}
    \label{fig:overview}
\end{figure*}

In this paper, we present a new, principled method for unsupervised learning of video representations from unlabeled video data. It is based on the observation that the optimized combination of multiple self-supervised tasks\footnote{In this work we use unsupervised and self-supervised interchangeably.}, which are additionally encouraged by multi-modal distillation, is often sufficient to learn good feature representations. Importantly, we demonstrate that such combination could be found without per-class or per-video labeling by instead matching the representation statistics to a general power distribution of video classes, e.g., Zipf's law~\cite{zipf1949human}.


Our approach is to train the network so that its intermediate representations reflect not just information directly obtained from its own input modality (e.g., RGB image) but also information from different modalities (e.g., grayscale, optical flow, and audio). The idea is that synchronized multi-modal data sources should benefit representation learning of each other as they correspond to the same content.
This is done by the introduction of `distillation' \cite{hinton2015distilling} losses between multiple streams of networks. The distillation losses, as well as the self-supervised tasks, do not rely on human annotation or supervision. 
As a result, our approach is formulated as a multi-modal, multi-task unsupervised learning, where the tasks include single-modality tasks like frame ordering as well as multi-modality tasks like video-audio alignment.

However, combining multiple different self-supervised task losses and distillation losses for unlabeled representation learning is a challenging problem, as
certain tasks and modalities are more relevant to the final task than others and different loss functions have different scales.
Thus, we newly introduce the concept of using an evolutionary algorithm to obtain a better multi-modal, multi-task loss function that appropriately combines all the losses to train the network. AutoML has successfully been applied to architecture search \cite{liu2018progressive} and data augmentation \cite{cubuk2018autoaugment}. Here we extend this concept to unsupervised learning by automatically finding the weighting of self-supervised tasks for video representation learning. 
The `fitness' of this evolution could naturally be measured with task specific labels (e.g., accuracy). However, we instead propose a purely unsupervised alternative based on power law distribution matching between the datasets, by using KL divergence constraints. These constraints do not require any labeled data, enabling fully unsupervised and unlabeled learning.

Our goal is to find video feature representations, based on a \textbf{single} RGB network, that can seamlessly improve supervised or unsupervised tasks without 
additional annotations. The main contributions are:

\begin{itemize}[nosep,leftmargin=*]
  \item Formulation of unsupervised learning as multi-modal, multi-task learning, including distillation tasks to transfer features across modalities into a single-stream network. Once learned, it allows for faster representation computation while still capturing multi-modal features.
  \item Evolutionary search for a loss function that automatically combines self-supervised and distillation tasks that are beneficial for unsupervised representation learning.
  \item Introduction of an unsupervised representation evaluation metric based on power law distribution matching, which requires no labels and performs similarly to the label-guided one.
\end{itemize}

This work makes the surprising finding that large amounts of unlabeled data, combined with self-supervised tasks and the power law distribution matching, produces very powerful feature representations which are only rivaled by large datasets with very extensive data labeling: Our feature representations (obtained with zero labels) outperform ImageNet pre-training, and pre-training on small and medium-size labeled video datasets; it is only outperformed by Kinetics pre-training with full annotations based on human labeling of more than 200,000 videos. 
Further, the proposed representations outperform Kinetics training, when fine-tuned with Kinetics labels. 
We refer to the model as \textbf{`ELo'}, as it is based on evolving unsupervised losses.

\section{Related Work}

\textbf{Unsupervised Video Representation Learning:} Obtaining labeled video data is expensive, and unlabeled video data is plentiful, and there have been many methods proposed for self-supervised learning of video representations. Some tasks take advantage of the temporal structure in videos, such as predicting if frames appear in order, reverse order, shuffled, color-consistency across frames, etc. \cite{fernando2017self, misra2016shuffle, isola2015learning, lee2017unsupervised,pickup2014seeing,wei2018learning,jayaraman2016slow,kim2018self,wang2017transitive,wang2019learning,vondrick2018tracking}. 
Other work has explored using the spatial structure present in images, such as predicting relative spatial location of image patches \cite{noroozi2016unsupervised} or tracking patches over time \cite{wang2015unsupervised}, showing promising results. Reconstruction or prediction of future frames\cite{srivastava2015unsupervised}, or time-contrastive learning\cite{timecontrastive,sermanet2017time} to obtain representations has also been successful. Learning representations taking advantage of audio and video features has been explored by predicting if an audio clip is from a video \cite{arandjelovic2017look} or if audio and video are temporally aligned \cite{owens2018audio,chung2016out,korbar2018cooperative,Arandjelovic2018objects}.

Multi-task self-supervised learning has also shown promising results \cite{doersch2017multi,ren2018cross,zamir2018taskonomy}, where tasks are assumed to have equal weights and are not multi-modal.
Generating weak labels using $k$-means clustering on CNN features \cite{bautista2016cliquecnn,caron2018deep} or using clustering with meta-learning \cite{hsu2018unsupervised} has also been explored.
In this paper, we propose a generalized approach to unsupervised representation learning, allowing for multi-modal inputs and automatic discovery of the tasks that benefit recognition performance.

\textbf{Activity Recognition:} Activity recognition is a active area of vision research, with a variety of methods proposed~\cite{wang2011action,tran2014c3d,simonyan2014two,feichtenhofer2018slowfast}. 
With the introduction of large activity recognition datasets (e.g., Kinetics and Moments in Time \cite{kay2017kinetics,monfort2018moments}), much more accurate deep video CNNs are possible \cite{carreira2017quo}. We here show that they can be further improved by unsupervised representation learning.

\section{Method}
We formulate unsupervised video representation learning as a combination of multi-task, multi-modal learning. The objective is not only to take advantage of multiple self-supervised tasks for the learning of a (good) representation space, but also to do so across multiple modalities. The idea is that models from synchronized multi-modal data, sharing the same semantic content, will benefit the representation learning of each other. We encourage that via introducing `distillation' losses. At the same time, each modality may have multiple self-supervised tasks with their corresponding losses. Fig. \ref{fig:overview} illustrates the multi-task, multi-modal formulation with multiple losses and distillation, Section~\ref{sec:multi} has the details.

To facilitate multi-task, multi-modal learning, importantly, we introduce the new concept of automatically \emph{evolving} the main loss function. Certain tasks and modalities are more relevant to the final task so the representation needs to focus on those more than the others. The idea is to computationally search for how different multi-task and distillation losses should be combined, instead of constructing a loss function by trial-and-error. We discuss this more in Section~\ref{sec:evolution}. 

One key technical question is how one can guide the evolution without a pre-defined task or a fitness function. We propose an unsupervised method to evaluate each loss function, based on matching of the power law distribution of activity classes (Section~\ref{sec:zipf}).


\begin{figure*}
    \centering
    \includegraphics[width=\linewidth]{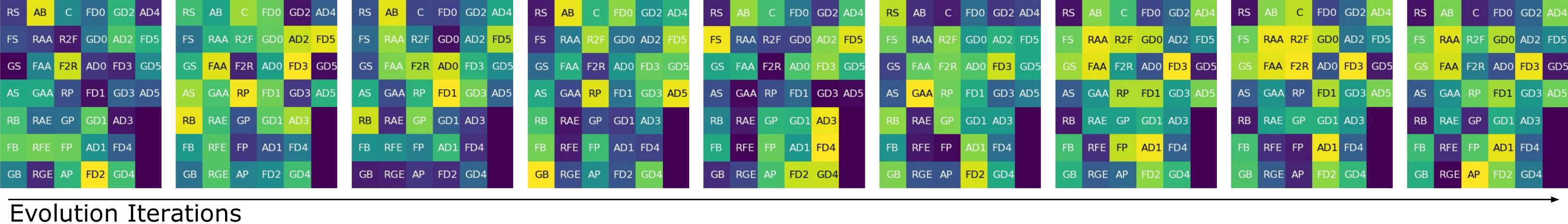}
    \caption{Evolution of the weights deciding our final loss function. Each square represents a $\lambda_{m,t}$ and how it changes over the evolutionary search. 
    The weight symbols are as follows: the first letter is representation modality (R=RGB, A=Audio, F=Flow, G=Grey), The tasks are S=Shuffle, C=colorize, A=Audio align, P=Future prediction, B=backward detection, D=Distill, E=Embed. The numbers indicate the layer the distillation loss is applied.}
    \label{fig:loss-fn-evol}
\end{figure*}

\subsection{Unsupervised multi-modal learning}
\label{sec:multi}

We construct a CNN for each modality. Each network is trained using several tasks not requiring labeled video data, and the information from each modality is combined using distillation \cite{hinton2015distilling} (Fig. \ref{fig:overview}). More specifically, we take advantage of multiple self-supervised tasks, such as frame reconstruction, future frame prediction, and frame temporal ordering (Section \ref{subsec:tasks} discusses them in detail). Each of these tasks will yield an unsupervised loss for training. Learning with multiple self-supervised tasks makes our representations more generic, as they need to generalize to many tasks and are more transferable to unseen tasks.

For each modality, $m$ and its input $I_m$, we build an embedding network, $E_m$, which generates an embedded representation of the input: $x_m = E_m(I_m)$. $x_m$ is the feature representation for modality $m$. Our embedding networks are (2+1)D ResNet-50 models, which take advantage of both 2D spatial convolutions and 1D temporal convolutions to represent videos; they provide state-of-the-art performance on video understanding tasks. As mentioned, for each modality, we consider several learning tasks, for example, frame reconstruction. Each of the tasks per modality has its own loss function. ${L}_{m,t}$ is the loss from task $t$ and modality $m$ and $\{t_1, t_2 ..., t_{N_m}\}$ is the set of tasks for the modality.

Further, to better take advantage of the multi-modal representations, we use distillation to `infuse' the other modalities into the RGB network at different locations. Our final objective is to train a single RGB network that provides a strong representation for video understanding. Our formulation allows the RGB network to learn representations from various tasks and modalities.

We combine the multi-task losses for unsupervised training, and per each modality, by a weighted sum, and we further combine it with a number of distillation losses $\mathcal{L}_d$ which fuse or synchronize the multiple modalities:
\begin{equation}
    \mathcal{L} = \sum_m \sum_t \lambda_{m,t} \mathcal{L}_{m,t} + \sum_d \lambda_d \mathcal{L}_d
    \label{eq:main}
\end{equation}
where $\lambda_{m,t}$ and $\lambda_d$ are the weights of the losses. The weighted sum $\mathcal{L}$ is the loss we use to train the entire model.

\subsubsection{Distillation}
Distillation was introduced to train smaller networks by matching representation of deeper ones \cite{hinton2015distilling}, or for generally transferring knowledge from pre-trained networks. Here, we use distillation to `infuse' representations of different modalities into the main, RGB network. Note that we distill representations jointly while training. The distillation losses learn features by transferring information across modalities. More specifically, our formulation allows for the distillation of audio, optical flow and temporal information into a single, RGB-based convolutional neural network. 

The distillation loss is the $L_2$ difference between the activations of a layer in the main network $M_i$ and a layer in another network $L_i$. Such constraint encourages the activations of the main network to match the activations of the other network, infusing other features into the main network.
\begin{equation}
    \mathcal{L}_{d}(L_i, M_i) = ||L_i - M_i||_2
\end{equation}

Distillation has previously been used for combining networks such as ensembles \cite{hinton2015distilling} or learning to predict optical flow features from RGB \cite{stroud2018d3d}, here we are extending its usage for multi-modal representation learning from unlabeled video data.
While in principle distillation can happen across all modalities, we do distillation only into the RGB stream, so as to obtain a final single-tower efficient representation for learning subsequent tasks.
Using the learned weights for the RGB network, we can then extract representations for a set of videos. 

\subsection{Evolving an unsupervised loss function}
\label{sec:evolution}

Our representation learning is governed by the weight coefficients of the loss in Equation~\ref{eq:main}, and they need to be appropriately determined. 
The weighting supposedly reflects the importance or relevance of each task and modality to the main task; for example the optical flow modality may be important for tracking, whereas audio may give more information for temporal segmentation of videos in certain settings. 

\subsubsection{Unsupervised loss construction}
Instead of manually constructing a loss function, we evolve the loss function by taking advantage of well-established evolutionary algorithms, e.g., ~\cite{goldberg91acomparative}. More specifically, our search space consists of all the weights of the loss function, both task weights and distillation weights. Each $\lambda_{m,t}$ or $\lambda_{d}$ is constrained to be in $[0,1]$. Our evolutionary algorithm maintains a pool (i.e., population) of individuals where each individual is a set of weight values that compose the final loss function.

\subsubsection{Unsupervised Zipf distribution matching}
\label{sec:zipf}

The evolutionary algorithm requires evaluation of the loss function (i.e., fitness measure) at each round to optimize the loss weight coefficients. We propose a new, unsupervised method for this. 
In order to measure the fitness of each individual (i.e., the set of weights to combine the tasks and modalities to form the final loss), we apply a  $k$-means clustering on the representation learned with the corresponding loss function, and analyze the cluster distributions. We first train the network using a smaller subset (100k) of unlabeled, random YouTube videos for 10,000 iterations (using the corresponding loss function). 
We then use a subset of random YouTube videos and similarly extract representations $x_{RGB} = E_{RGB}(I)$. Given these representations, we can then cluster them into $k$ clusters.

$k$-means clustering can be viewed as a Gaussian Mixture Model with fixed variances and we calculate probabilities of each feature vector belonging to a cluster, which reduces to computing the distance. Specifically, for the cluster centroids $\{c_1, c_2, \ldots c_k\}$ where $c_i \in \mathcal{R}^D$ (a $D$-dimensional vector), we can compute the probability of a feature vector $x\in \mathcal{R}^D$ belonging to a cluster $c_i$ as:
\begin{equation}
    p(x | c_i) = \frac{1}{\sqrt{2\sigma^2\pi}}\exp{\left(-\frac{(x - c_i)^2}{2\sigma^2}\right)}
\end{equation}
Since we are (naively) assuming all clusters have the same variance (for simplicity, let $2\sigma^2 = 1$) and an equal prior over all clusters, we can use Bayes rules to rewrite as:
\begin{equation}
\begin{split}
    p(c_i | x) &= \frac{p(c_i) p(x | c_i)}{\sum_j^k p(c_j)p(x|c_j)} = \frac{\exp{-\frac{(x - c_i)^2}{2\sigma^2}}}{\sum_{j=1}^k\exp{-\frac{(x - c_j)^2}{2\sigma^2}}}\\ &= \frac{\exp{-(x - c_i)^2}}{\sum_{j=1}^k\exp{-(x - c_j)^2}}
\end{split}
\end{equation}
which we note is the standard softmax function applied to the squared distances from a feature $x$ to a cluster center $c_i$.

As observed in many large activity recognition datasets, like AVA~\cite{ava2017} and Kinetics~\cite{kay2017kinetics}, the activity classes of videos follow a Zipf distribution. We can use this as a prior constraint on the distribution of the videos in these clusters. Specifically, given the above probability of each video belonging to each cluster, and the Zipf distribution, we compute the prior probability of each class as $q(c_i) = \frac{1/i^s}{H_{k,s}}$ where $H$ is the $k$th harmonic number and $s$ is some real constant. We then let $p(c_i) = \frac{1}{N}\sum_{x\in V} p(c_i | x)$, the average over all videos in the set. Using these two probability functions representing the classes/clusters, we can minimize the KL divergence:
\begin{equation}
\label{eq:kl}
    KL(p||q) = \sum_{i=1}^k p(c_i)\log{\left(\frac{p(c_i)}{q(c_i)}\right)}
\end{equation}
By using this as a fitness metric, it poses a prior constraint over the distribution of (learned) video representations in clusters to follow the Zipf distribution. Note that this method requires no labeled data and is fully unsupervised. We refer to this entirely unsupervised method as `ELo'.

\paragraph{Weakly-supervised baseline:} As an upper-bound and an alternative to our approach, we use a handful of class labels to evaluate the clustering (which is referred to as ELo-weak). This is done for the sake of comparison and is also a good alternative to align the final loss to a downstream video classification task.
A subset of HMDB is used, and $k$-means clustering is applied to the output representation of the RGB stream. The clusters are used for nearest-neighbors classification and the accuracy is the fitness of the individual.  
Due to randomness in $k$-means clustering, in both settings, we run this processes 20 times and average fitness across all trials.


\subsubsection{Loss evolution}

As is typical with evolution approaches, the evolution of the loss is driven by mutations.
Since our search space consists of continuous values in $[0,1]$, we compare two different evolutionary strategies: tournament selection \cite{goldberg91acomparative} and CMA-ES \cite{cmaes}. For the tournament selection search, we mutate an individual loss function by randomly picking one weight and assigning new value in uniformly sampled from $[0,1]$. For CMA-ES, at each iteration, all the components are changed based on the fitness of all the individuals in the evolution pool. For tournament selection, we evolve the loss function for 2000 rounds, generating and evaluating 2000 different loss functions and we use 250 rounds with CMA-ES, finding much faster convergence. Fig.~\ref{fig:loss-fn-evol} shows an example how our weights evolve over the rounds and Table \ref{tab:es} compares the performance of different search methods.
Since everything is differentiable, we could learn these weights also with gradient descent, however, we leave this for future exploration as taking the derivative of the entire network w.r.t. to task weights is non-trivial.

\begin{figure}
    \centering
    \includegraphics[width=0.85\linewidth]{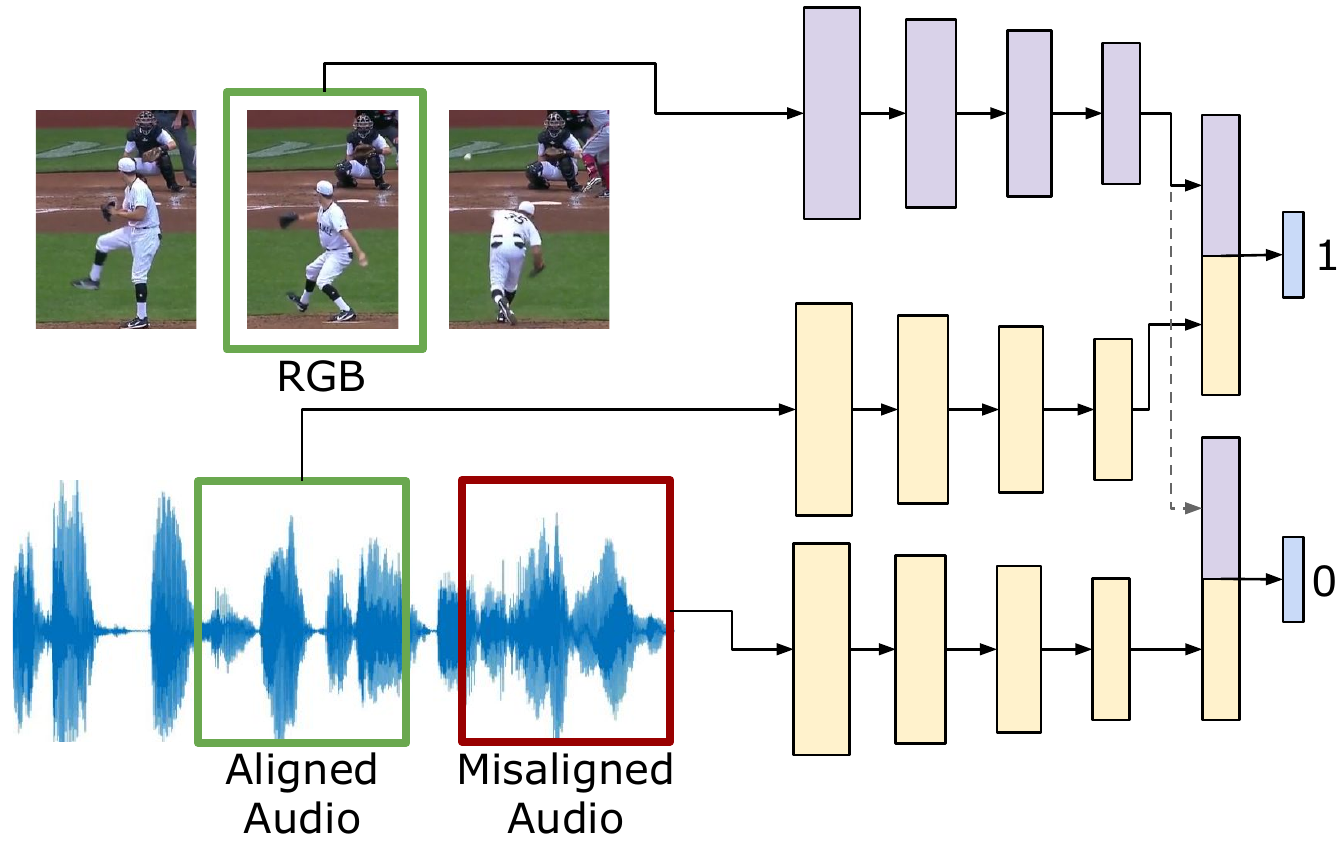}
    \caption{Example of the multi-modal alignment tasks. The networks take input of temporally aligned RGB and Audio (or other modalities) and a sample from one modality that is temporally different. The network is trained to predict if a pair is temporally aligned or not.}
    \label{fig:align-tasks}
\end{figure}

\subsection{Self-supervised tasks}
\label{subsec:tasks}

Many tasks have been designed for unsupervised learning. We describe briefly the tasks we employ for our representation learning. Importantly, we allow many possible tasks and let the evolved loss function automatically discover which tasks are important and the optimal relative weightings. We also use tasks like DeepCluster \cite{caron2018deep} and local aggregation \cite{zhuang2019local}.

\textbf{Reconstruction and prediction tasks:} Given the representation for a modality, $x_m$, we use a decoder CNN to generate the output. As the reconstruction is only used as a supervision signal, we do not need to generate extremely high quality reconstructions. Thus, we use a small, cheap decoder with only 6 convolutional layers, saving both memory and training time. Once the unsupervised learning is complete, we discard the decoders. Further, following previous work \cite{stroud2018d3d}, our decoders have no temporal convolution, forcing the representations to contain all needed temporal information, which is desirable for video understanding.
Each modality (e.g., RGB, optical flow, and audio) will be reconstructed. We also use several cross-modality transfer tasks: RGB to Flow, Flow to RGB, etc. 

Another way to learn video temporal structure is to train a decoder to predict the next $N$ frames given $T$ frames. Future prediction of frames has previously been used for representation learning \cite{srivastava2015unsupervised} and we perform this task for each modality.
For these tasks, we minimize the $L_2$ distance between the ground truth ($I$) and predicted output ($\hat{I}$):
\begin{equation}
    \mathcal{L}_R(\hat{I}, I) = ||\hat{I}-I||_2
\end{equation}

\textbf{Temporal ordering:} We use two tasks to learn representations that take advantage of temporal structure: binary classification of ordered frames and shuffled frames \cite{misra2016shuffle} and binary classification of forward and backward videos \cite{pickup2014seeing}. We use a single, fully-connected layer to make a binary classification of the representation: $p = W x_m$ ($x_m$ is the representation for a modality).
These are trained to minimize binary cross-entropy:
\begin{equation}
    \mathcal{L}_B(p, y) = -(y\log(p) + (1-y)\log(1-p))
\end{equation}
where $p$ is the output of the binary classifier and $y$ is the ground truth.


\textbf{Multi-modal contrastive loss:} As videos contain multiple modalities, we want to take advantage of these different representations to learn an a generic representation by using a multi-modal embedding space. Given the representations for each modality, $x_m$, we minimize a contrastive loss between various embedding spaces:
\begin{equation}
    \mathcal{L}_c (x_{1}, x_{2}, x_n) = ||x_{1}-x_{2}||_2 + \max(0, \alpha - ||x_{1} - x_n||_2)
\end{equation}
where $x_{m1}$ and $x_{m2}$ are representations from the same video but different modalities and $x_n$ is a representation from a different video.
This task encourages the representations from the same video, but different modalities, to be close in the representation space, while representations from different videos are further apart.

\textbf{Multi-modal alignment:} We can further take advantage of both the temporal information and the multi-modal data by performing a multi-modal alignment task, illustrated in Fig. \ref{fig:align-tasks}. The networks take input of temporally aligned samples from two modalities, and a sample from one modalitiy from a temporally different region. The model is trained to make a binary prediction if the two samples are temporally aligned.

\begin{table}
\small
    \centering
    \begin{tabular}{lccc}
    \toprule
       Method  & $k$-means & 1-layer & fine-tune \\
    \midrule
        \multicolumn{4}{l}{\textbf{Supervised using additional labeled data}}\\
        Scratch (No Pretraining) & 15.7 & 17.8 & 35.2 \\
        ImageNet Pretrained & 32.5 & 37.8 & 49.8 \\
        Kinetics Pretrained & 68.8 & 71.5 & 74.3 \\
        \midrule
        \multicolumn{4}{l}{\textbf{Unsupervised using unlabeled videos}}\\
        Frame Shuffle \cite{misra2016shuffle} & 22.3 & 24.3 & 28.4 \\
        Reverse Detection \cite{pickup2014seeing} & 21.3 & 24.3 & 27.5\\
        Audio/RGB Align \cite{owens2018audio,korbar2018cooperative}     & 32.4 & 36.8 & 40.2 \\
        RGB to Flow   & 31.5 & 36.4 & 39.9 \\
        Predicting 4 future frames & 31.8 &  35.8 & 39.2 \\
        Joint Embedding & 29.4 & 32.5 & 38.4 \\
        \midrule
        \multicolumn{4}{l}{\textbf{Ours, weakly-sup clustering, using unlabeled videos}}\\
        Evolved Loss - ELo-weak & 45.7 & 64.3 & 67.8 \\
        \midrule
        \multicolumn{4}{l}{\textbf{Ours, unsupervised, using unlabeled videos}}\\
         Random Loss (unsup.)  & 26.4  & 26.9  & 31.2  \\
        Evolved Loss - ELo (unsup.) & 43.4 & 64.5 & 67.4 \\
    \bottomrule
    \end{tabular}
    \caption{Evaluation of various self-supervised methods on HMDB51 \cite{kuehne2011hmdb}. We compare to a randomly initialized, ImageNet pretrained and Kinetics pretrained networks. We also compare to various single-task baselines, an average of 10 randomly sampled loss functions, and the evolved loss function using both fitness metrics. All tasks were trained on our random, unlabeled YouTube videos.}
    \label{tab:hmdb-1}
\end{table}

\section{Experiments}

\subsection{Datasets} 

\paragraph{Unsupervised data source.} We use two million random, unlabeled YouTube video clips sampled randomly from the Youtube-8M dataset~\cite{abu2016youtube} (limiting the size for computational reasons). Previous works on self-supervised learning used videos from datasets (e.g., Kinetics or AudioSet in \cite{korbar2018cooperative}) while ignoring the labels, leading a bias in the dataset, as those videos are trimmed to intervals with specific activities. Using a random sample from Youtube is less prone to bias as the videos are user generated, the labels are automatically tagged (no human verification), 
and potentially offers a very large set for training (up to 8M). 
We have verified there is no overlap between those datasets and the ones the models are evaluated on (e.g., HMDB).



\textbf{Evaluation datasets and protocols.}
We used the following widely used datasets for evaluation: 
\textbf{HMDB~\cite{kuehne2011hmdb}},
\textbf{UCF101~\cite{UCF101}},
\textbf{Kinetics~\cite{kay2017kinetics}}.
We also used Imagenet~\cite{imagenet_cvpr09} and Kinetics for reporting results with pre-training, as is customary in previous work.
We use the standard protocols when evaluating video classification results on the labeled datasets, adopted by prior work as well. Please see the sup. material for dataset details.


\textbf{Implementation details.} We use a (2+1)D ResNet-50 as our backbone network. Given a loss function, we train the network for 100 epochs on 2 million unlabeled videos. The learning rate is set to 0.1 (during both the evolution and the final training). We use cosine learning rate decay with a warmup period of 2 epochs.

During search, we used smaller networks, similar to a ResNet-18 for faster learning. For search, the fitness of each model can be found in 4 hours using 8 GPUs. The final model uses 64 GPUs for 3 days to train (equivalent time to training I3D/(2+1)D ResNet-50 on Kinetics). 


\begin{table}
    \centering
    \begin{tabular}{lcc}
    \toprule
    Method & HMDB & UCF101 \\
    \midrule
    \multicolumn{3}{l}{\textbf{Supervised}}\\
    (2+1)D ResNet-50 Scratch   & 35.2 & 63.1 \\
    (2+1)D ResNet-50 ImageNet & 49.8 & 84.5 \\
    (2+1)D ResNet-50 Kinetics & 74.3 & 95.1 \\
    \midrule
    \multicolumn{3}{l}{\textbf{Unsupervised}}\\
    Shuffle \cite{misra2016shuffle} & 18.1 & 50.2\\
    O3N \cite{fernando2017self}  & 32.5 & 60.3 \\
    OPN \cite{lee2017unsupervised} & 37.5 & 37.5 \\
    Patch \cite{wang2015unsupervised} & - & 41.5 \\
    Multisensory \cite{owens2018audio} & - & 82.1 \\
    AVTS \cite{korbar2018cooperative} & 61.6 & 89.0 \\
    \midrule
    \multicolumn{3}{l}{\textbf{Weakly guided, HMDB}}\\
    Evolved Loss (ours) & 67.8 & 94.1 \\
    \midrule
    \multicolumn{3}{l}{\textbf{Unsupervised}}\\
    Evolved Loss (ours, no distiliation) &53.7 &84.2 \\
    Evolved Loss - ELo (ours) & 67.4 & 93.8 \\
    \bottomrule
    \end{tabular}
    \caption{Comparison to the state-of-art on HMDB51 and UCF101.  Note that previous approaches train on activity recognition datasets (e.g., Kinetics) are much more aligned to the final task, whereas we use random video clips. Even using more difficult data, we outperform the previous methods.
    (The top portion shows results for (2+1)D ResNet-50 with supervised pretraining as in Table~\ref{tab:hmdb-1}).}
    \label{tab:state-of-art}
\end{table}

\begin{figure}
    \centering
    \includegraphics[width=0.9\linewidth]{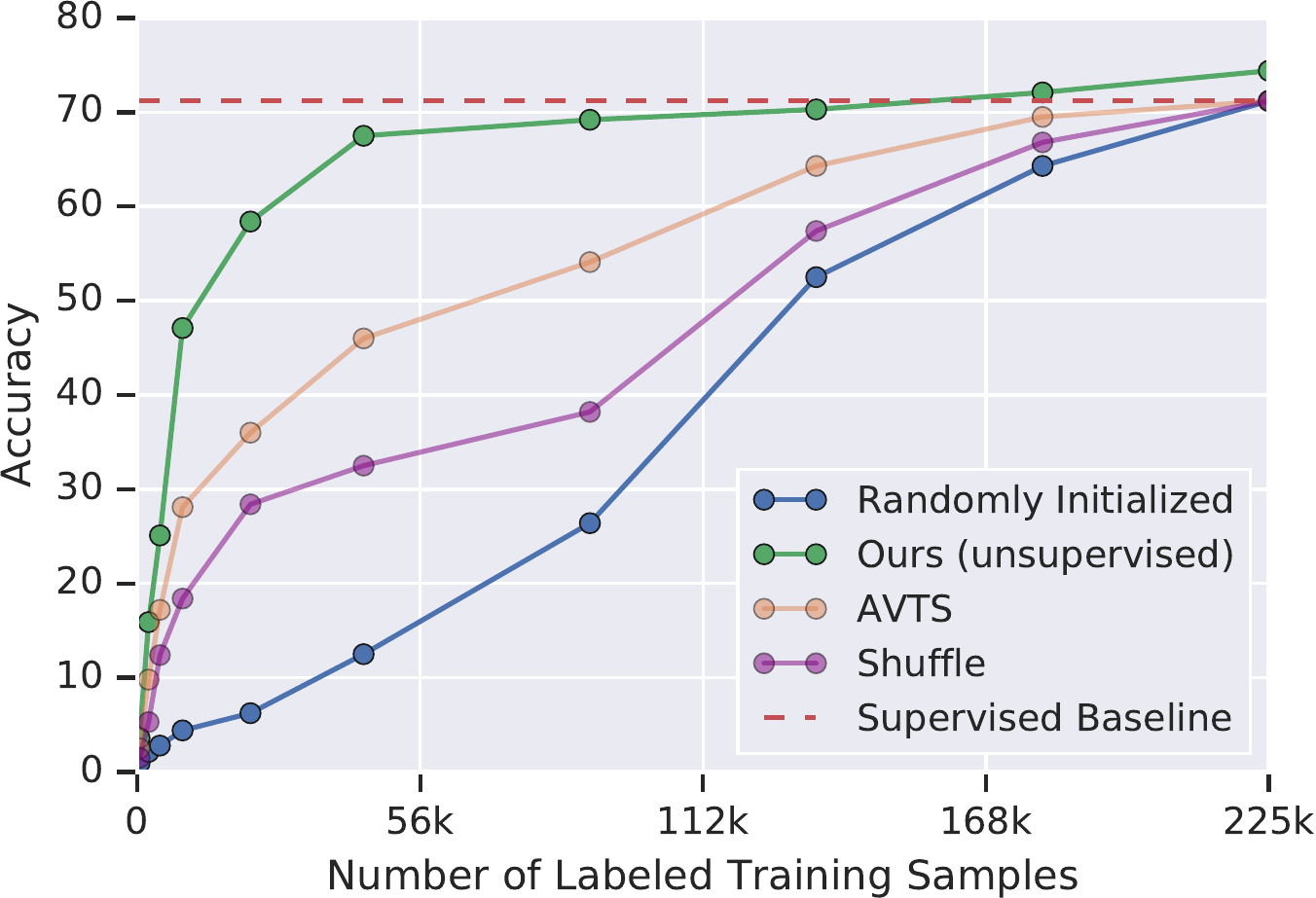}
    \caption{How much labeled, supervised data is needed once the unsupervised representation is learned. We achieve comparable performance with roughly half the data and outperform the supervised baselines using the entire dataset.}
    \label{fig:amounts-of-labeled-data}
\end{figure}

\subsection{Comparison to previous methods}

We evaluate our method in comparison to prior unsupervised and supervised representation learning. Specifically, we evaluate the representations in 3 settings: (1) $k$-means clustering of the representations (2) fixing the weights of the network and training a single, fully-connected layer for classification and (3) fine-tuning the entire network. These three evaluations are done by directly evaluating the representation as well as finetuning the entire network.

We find that while all approaches outperform the randomly initialized networks, only our evolved loss function outperforms ImageNet pretraining and performs comparably to the pretrained network with labeled Kinetics data (Table \ref{tab:hmdb-1}). Furthermore, we outperform all prior unsupervised methods. Our approach performs similarly to the weakly supervised version of our evolution method, despite being unsupervised. 
We also compare to a loss function randomly sampled from our search space, which performs poorly. We find that some tasks are not beneficial to representation learning, thus the evolution is quite important as this allows automatically finding the best tasks and weightings.

In Table \ref{tab:state-of-art}, we compare our approach to previous reported methods. 
We find that even though our approach is using more difficult unlabeled data, we still outperform the exiting methods by a significant margin.

We also find that distillation is extremely important. Without it, the RGB network can only take advantage of the other modalities through a limited number of tasks (e.g., RGB to Flow, audio/rgb alignment tasks). To learn a single RGB network, many important and high performing self-supervised tasks, such as flow, shuffling, can only influence the weights through distillation.

\begin{table*}
    \centering
    \small
    \begin{tabular}{l|cccccccccc}
    \toprule
     & \multicolumn{10}{c}{Number of Labeled Samples}\\
        Method & 400 & 2k & 4k & 8k & 20k & 40k & 80k & 120k & 160k & \shortstack{225k\\(all samples)}  \\
    \midrule
        Random Init &  0.93 & 2.1 & 2.8 & 4.4 & 6.2 & 12.5 & 26.4 & 52.5 & 64.3 & 71.2 \\
    \midrule
        Frame Shuffle & 1.5 & 5.3 & 12.4 & 18.4 & 28.4 & 32.5 & 38.2 & 57.4 & 66.8 & 70.9 \\
        Audio Align & 2.5 & 9.8 & 17.2 & 28.1 & 36.0 & 46.0 & 54.1 & 64.3 & 69.5 & 71.5 \\
        ELo (unsupervised) & 3.6 & 15.8 & 24.8 & 47.0 & 58.3 & 67.5 & 69.2 & \textbf{70.2} & \textbf{72.2} & \textbf{74.4}\\
    \bottomrule
    \end{tabular}
    \caption{Using different amounts of labeled samples on the currently available (March 2019) Kinetics-400 dataset using a (2+1)D ResNet-50. We achieve similar performance with only $\sim$50\% of the data. Using the entire dataset, we outperform the randomly initialized network.}
    \label{tab:num-labeled-samples}
\end{table*}

\begin{figure}
    \centering
    \includegraphics[width=0.49\linewidth]{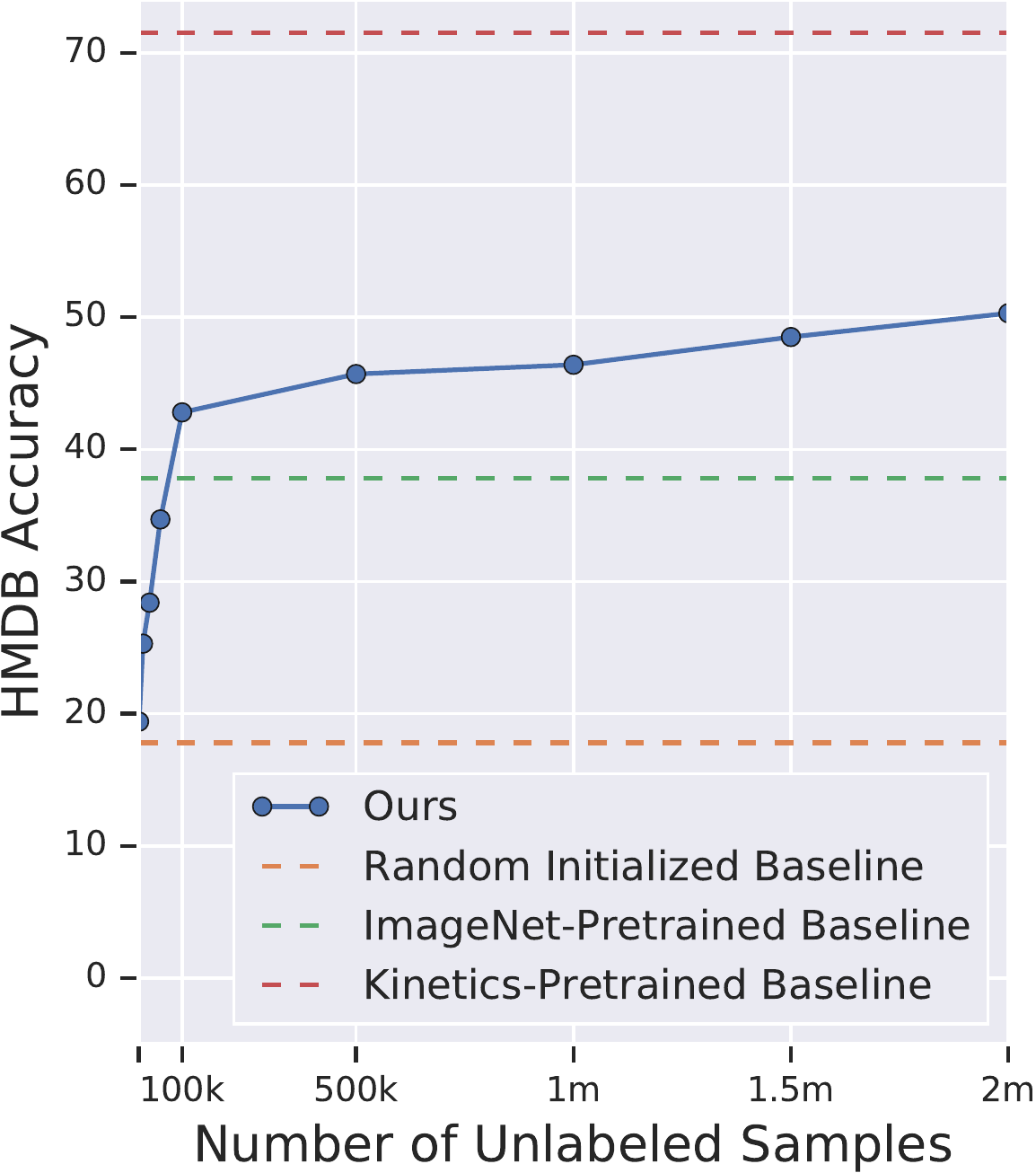} \includegraphics[width=0.49\linewidth]{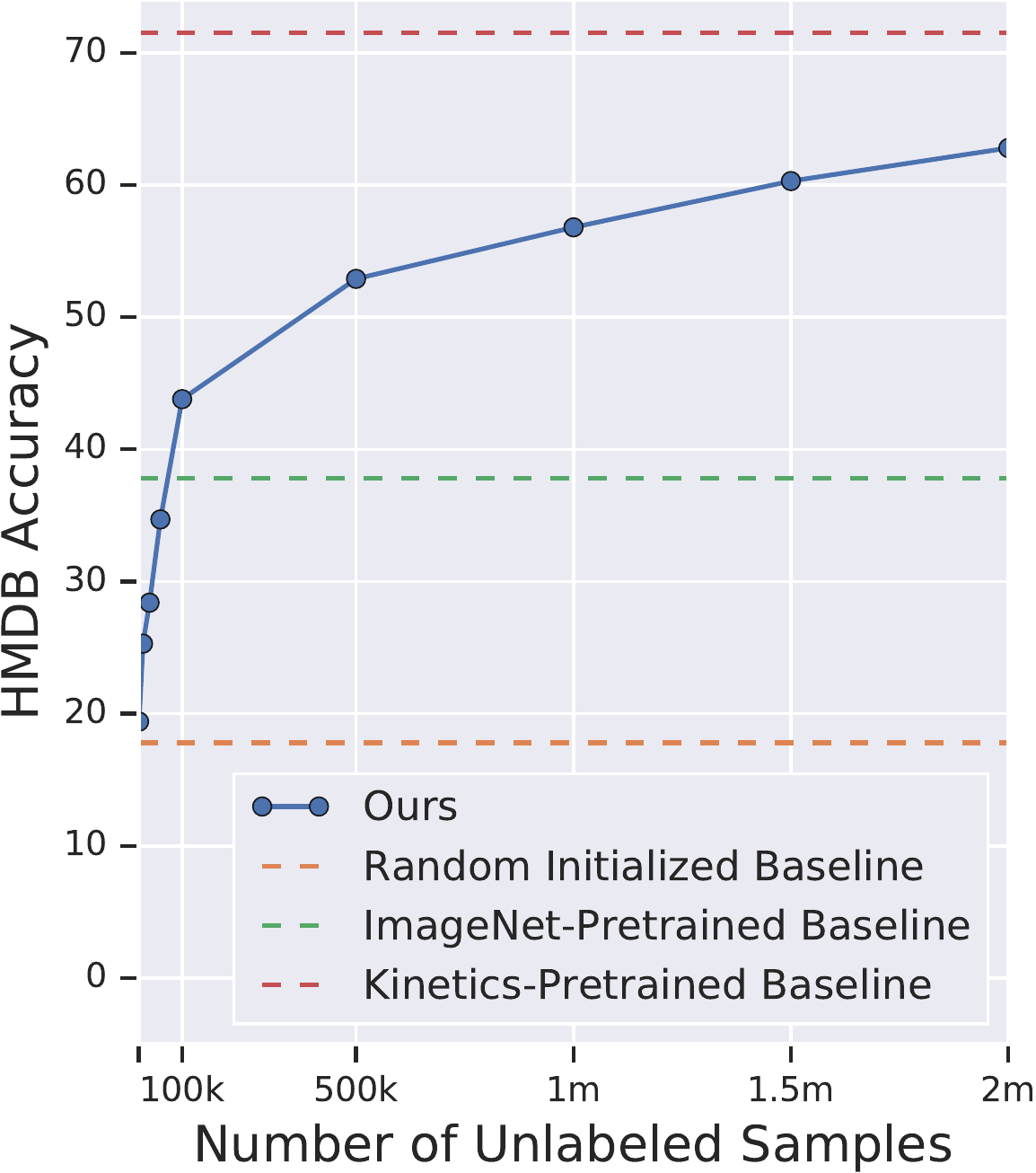}
    \caption{Comparisons of different amounts of unsupervised data. \textbf{Left}: Total number of training iterations fixed (i.e., less epochs as data is added). \textbf{Right}: Total number of epochs fixed (i.e., more iterations as more data is added). We observe that adding more data without increasing training time improves performance, while training longer on more data is better. On HMDB.}
    \label{fig:amounts-of-unsupervised-data}
\end{figure}

\begin{figure}
    \centering
    \includegraphics[width=1.1\linewidth]{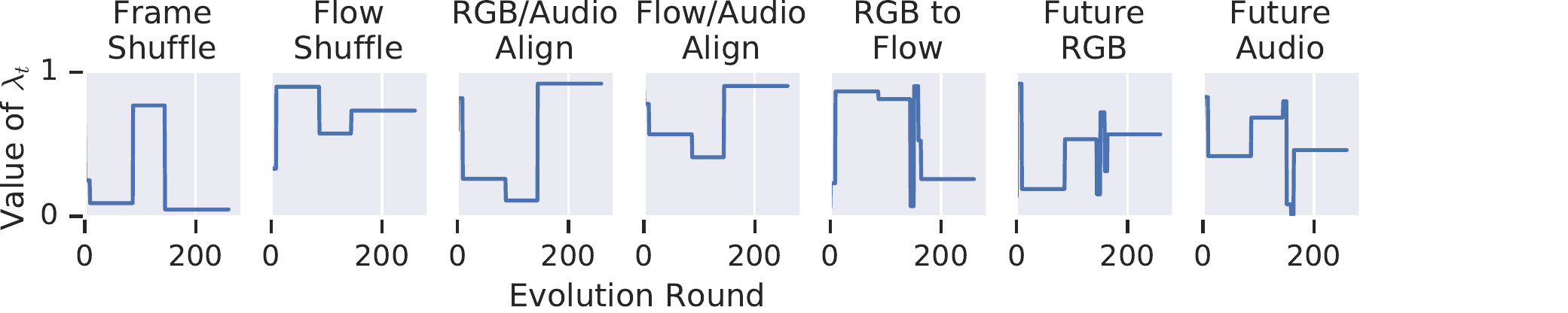}
    \caption{The values of the loss function for the various tasks throughout evolution. Higher weight values indicate the task is more important. The learned loss functions automatically finds the tasks that most benefit recognition.}
    \label{fig:tasks-evol}
\end{figure}

\begin{figure}
\vspace{-4mm}
    \centering
    \includegraphics[width=0.7\linewidth]{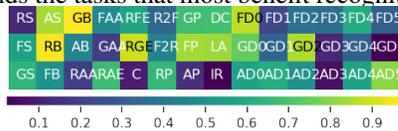}
    \caption{Heatmap visualization of the learned loss function. Higher values indicate the components importance. See Fig.~\ref{fig:loss-fn-evol} for description.}
    \label{fig:loss-heatmap}
    \vspace{-2mm}
\end{figure}

\subsection{Improving supervised learning}
Once we have learned a representation space using large amounts of unlabeled data, we want to determine how much labeled data is needed to achieve competitive performance. In Fig. \ref{fig:amounts-of-labeled-data} and Table \ref{tab:num-labeled-samples}, we compare various approaches trained using our unlabeled videos then fine-tuned on Kinetics using different amounts of labeled data. The Kinetics dataset has 225k labeled samples and we find that using only 25k (10\%) yields reasonable performance (58.1\% accuracy), only 11\% lower than our baseline, fully-supervised model using all samples.

We are able to match performance using only 120k samples, about half the dataset. Using the entire dataset, we outperform the baseline network, due to better initilizations and the distillation of modalities into the RGB stream.


\subsection{Benefit of additional unlabeled data}
We explore the effect of using different amounts of unlabeled data. Given a loss function, we train a network using $N$ unlabeled samples. As adding more data while keeping the number of epochs fixed increases the number of iterations, we compare the training both keeping the iterations fixed and the number of epochs fixed.

The results on HMDB are shown in Fig. \ref{fig:amounts-of-unsupervised-data}. When fixing the number of iterations to 100k, the performance increases as we add more data, even though the number of epochs (e.g., number of times each sample is seen) decreases. This suggests that during unsupervised training, the use of more, diverse data is beneficial, even when samples are seen fewer times. When fixing the number of epochs to 100, we find that adding more data further improves performance, suggesting that more training plus more data is best.

\begin{table}
\vspace{-4mm}
    \centering
    \small
    \begin{tabular}{c|cc}
    \toprule
        Method & Num iter. & Acc \\
        \midrule
        Random Search & 2000 & 52.4 \\
        Grid Search & 2000 & 57.3 \\
        Tournament Selection & 2000 & 61.4 \\
        CMA-ES & 250 & 67.4 \\
        \bottomrule
    \end{tabular}
    \caption{Comparison of best loss found with different evolutionary strategies evaluated on HMDB.}
    \label{tab:es}
\end{table}


\vspace{-2mm}
\subsection{Additional Analysis}

Examining the weights of the evolved loss function, $\lambda_{m,t}$ and $\lambda_d$, allows us to check which tasks are more important for the target task. Fig. \ref{fig:tasks-evol} illustrates the weights for several tasks ($\lambda_{m,t}$) over the 250 evolution rounds. We observe tasks such as RGB frame shuffle get very low weights, suggesting they are not very useful for the action recognition task. Tasks such as audio alignment are quite important. 
 The final fully-evolved loss is shown in Fig. \ref{fig:loss-heatmap}.

\begin{figure}
    \centering
    \vspace{-3mm}
    \includegraphics[width=0.48\linewidth]{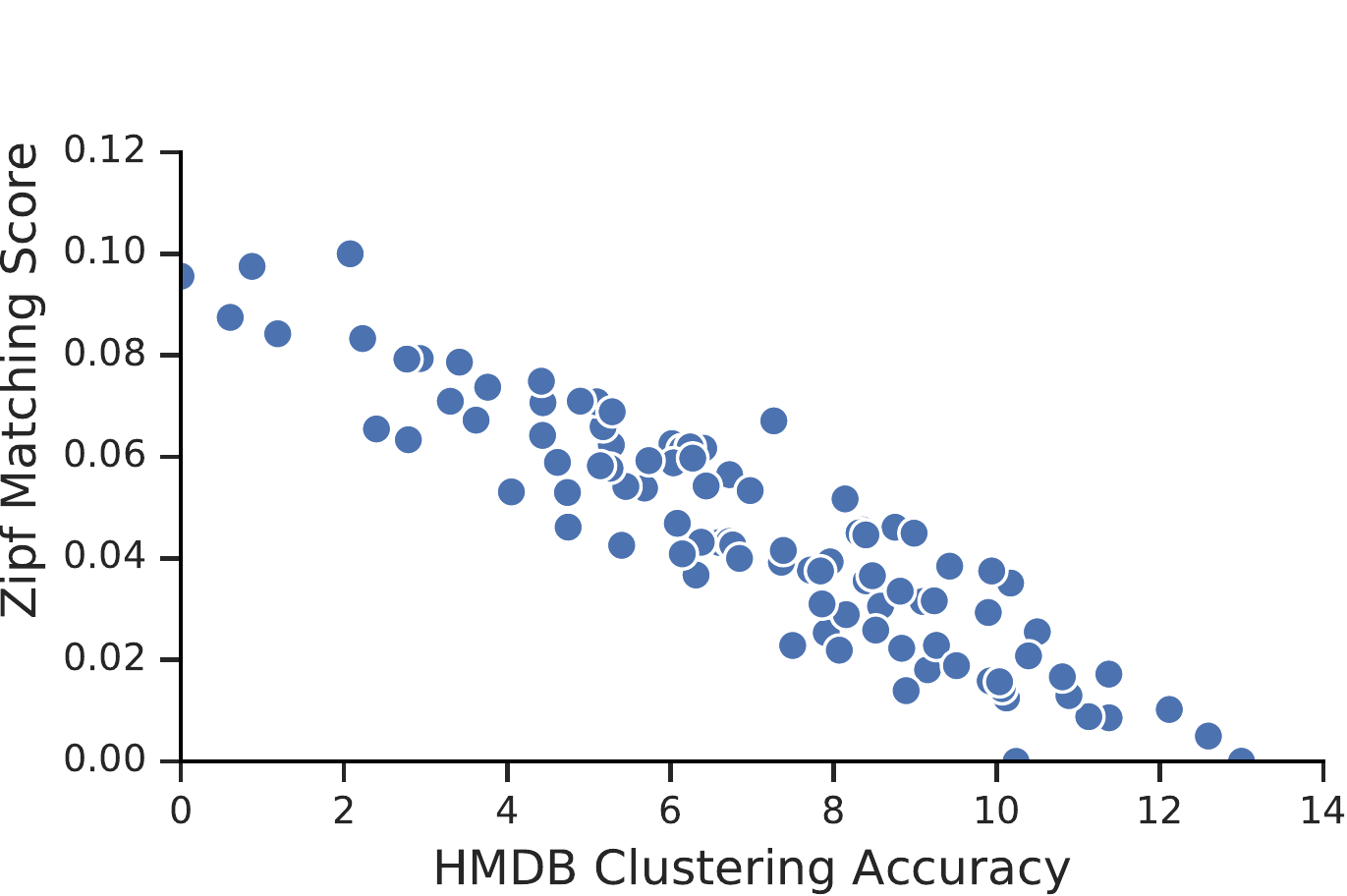}
    \includegraphics[width=0.4\linewidth]{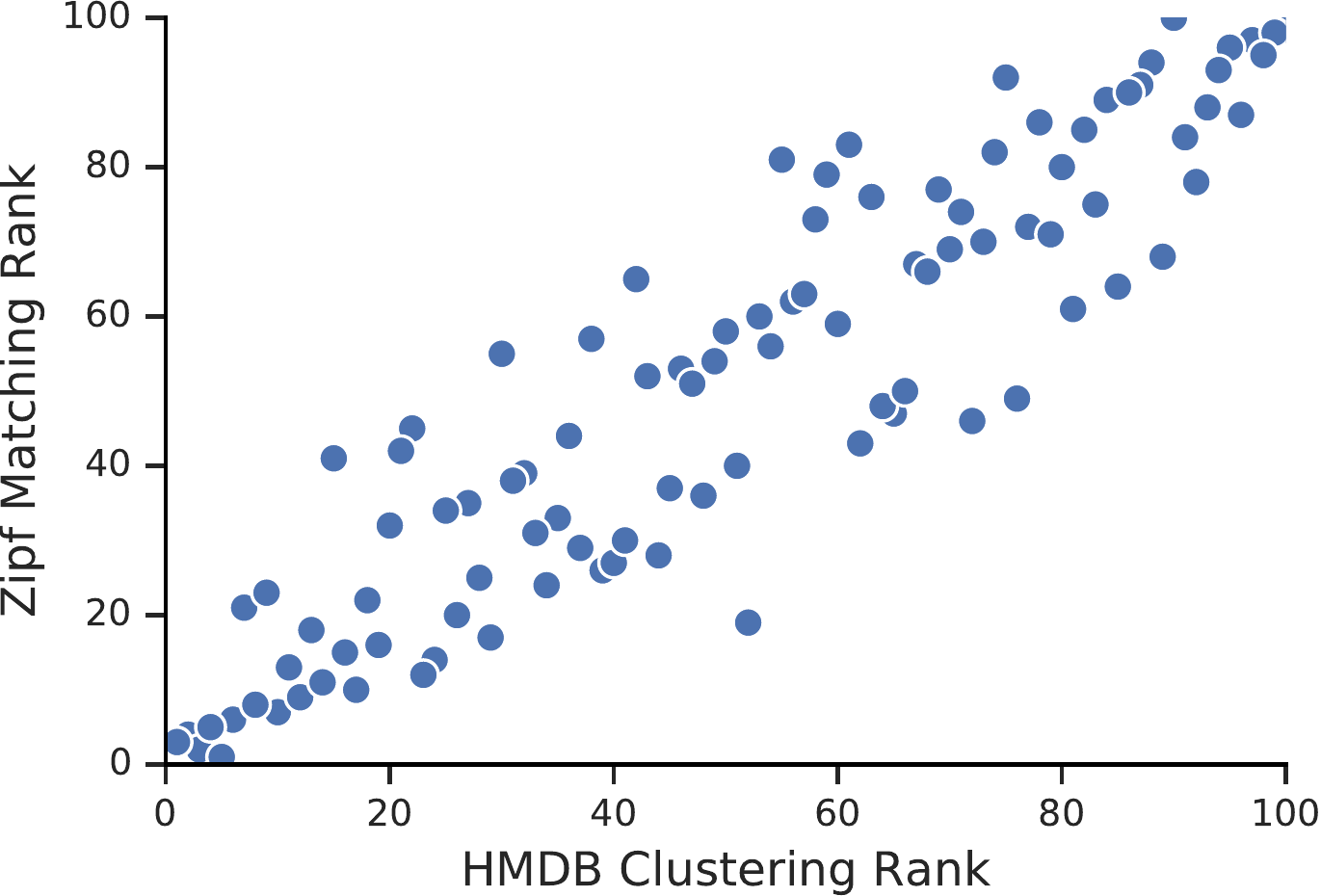}
    \caption{Comparison of the two fitness measures for 100 different loss functions. The plots show HMDB clustering and KL-divergence to Zipf distribution on random videos. The measures are quite correlated. \textbf{Left}: Fitness value (correlation $r=0.93$). \textbf{Right:} Ranking of each loss function (Spearmans's rho $\rho=0.91$).}
    \label{fig:hmdb-vs-kl}
\end{figure}

Table~\ref{tab:es} compares different search methods. As seen CMA-ES converges the most quickly and to the best fitness. In Fig. \ref{fig:hmdb-vs-kl}, we compare the two different fitness measures, finding strong correlation. This suggests that Zipf matching is suitable for unsupervised representation evaluation.

\vspace{-2mm}
\section{Conclusion}
\vspace{-1mm}
We proposed a unified framework for multi-task, multi-modal unsupervised video representation learning and found it benefits recognition tasks. We further introduced the concept of loss function evolution to automatically find the weights of the self-supervised tasks and modalities, with unsupervised fitness measure. 
We find powerful unsupervised video representations that outperform prior self-supervised tasks and can match or improve the performance of networks trained on supervised data.


{\small
\bibliographystyle{ieee_fullname}
\bibliography{egbib}
}

\clearpage
\newpage
\appendix

\begin{figure*}
\begin{subfigure}{.5\textwidth}
  \centering
  \includegraphics[width=.96\linewidth]{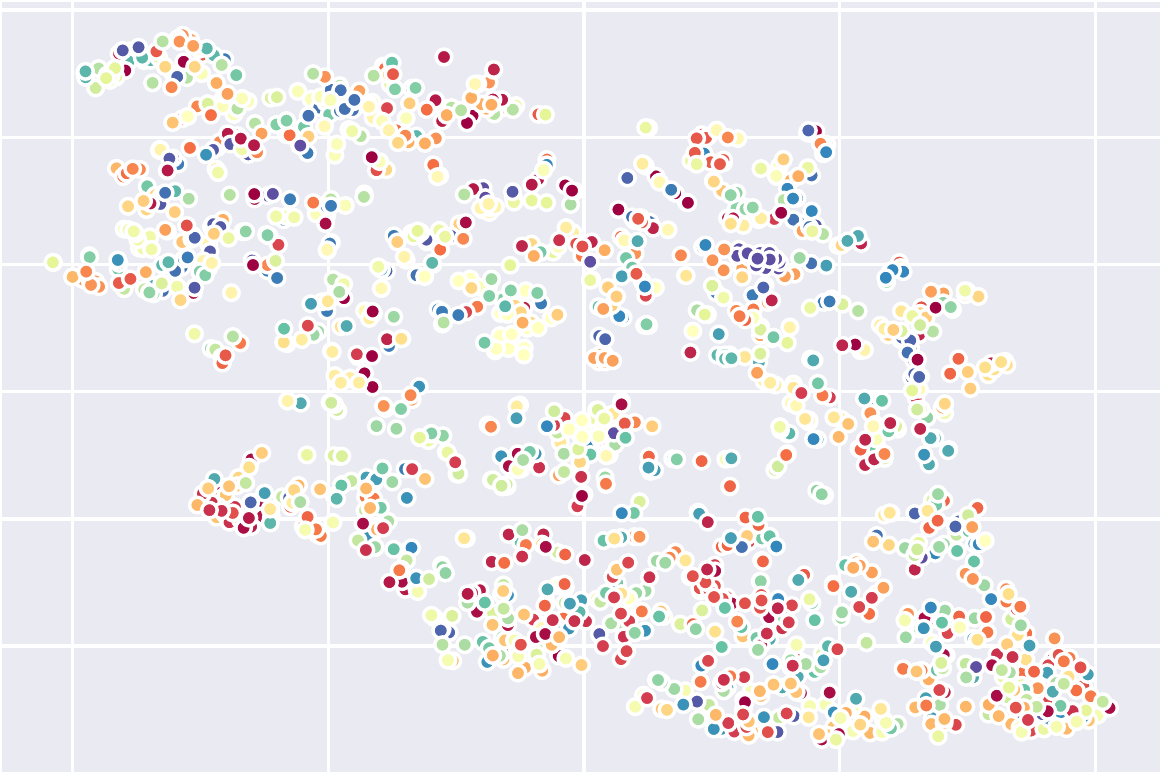}
  \caption{}
  \label{fig:tsnesfig1}
\end{subfigure}%
\begin{subfigure}{.5\textwidth}
  \centering
  \includegraphics[width=.96\linewidth]{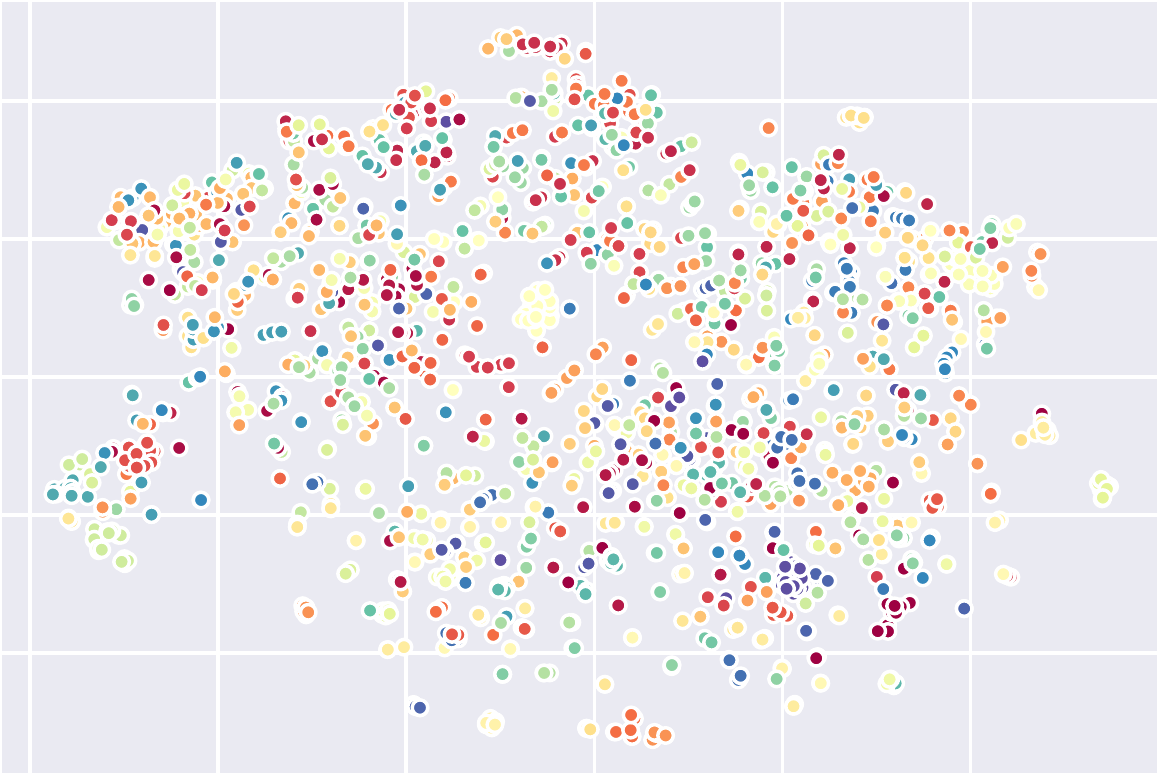}
  \caption{}
  \label{fig:tsnesfig2}
\end{subfigure}
\begin{subfigure}{.5\textwidth}
  \centering
  \includegraphics[width=.96\linewidth]{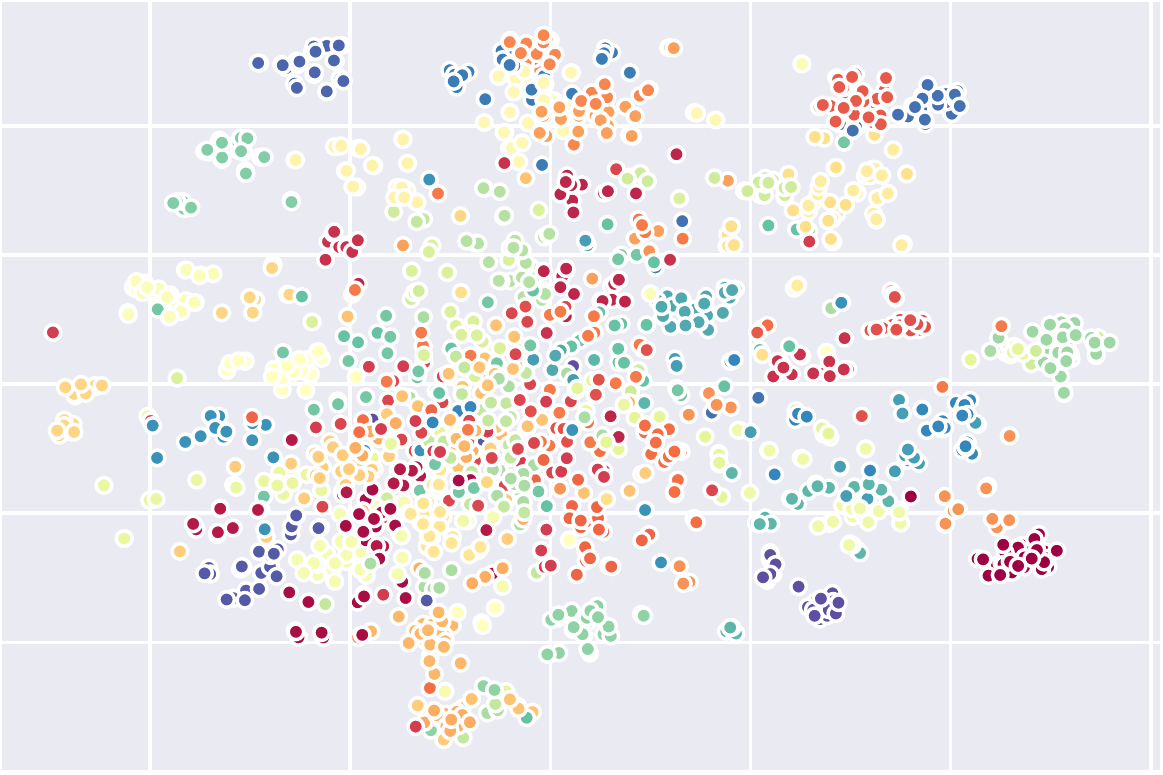}
  \caption{}
  \label{fig:tsnesfig3}
\end{subfigure}%
\begin{subfigure}{.5\textwidth}
  \centering
  \includegraphics[width=.96\linewidth]{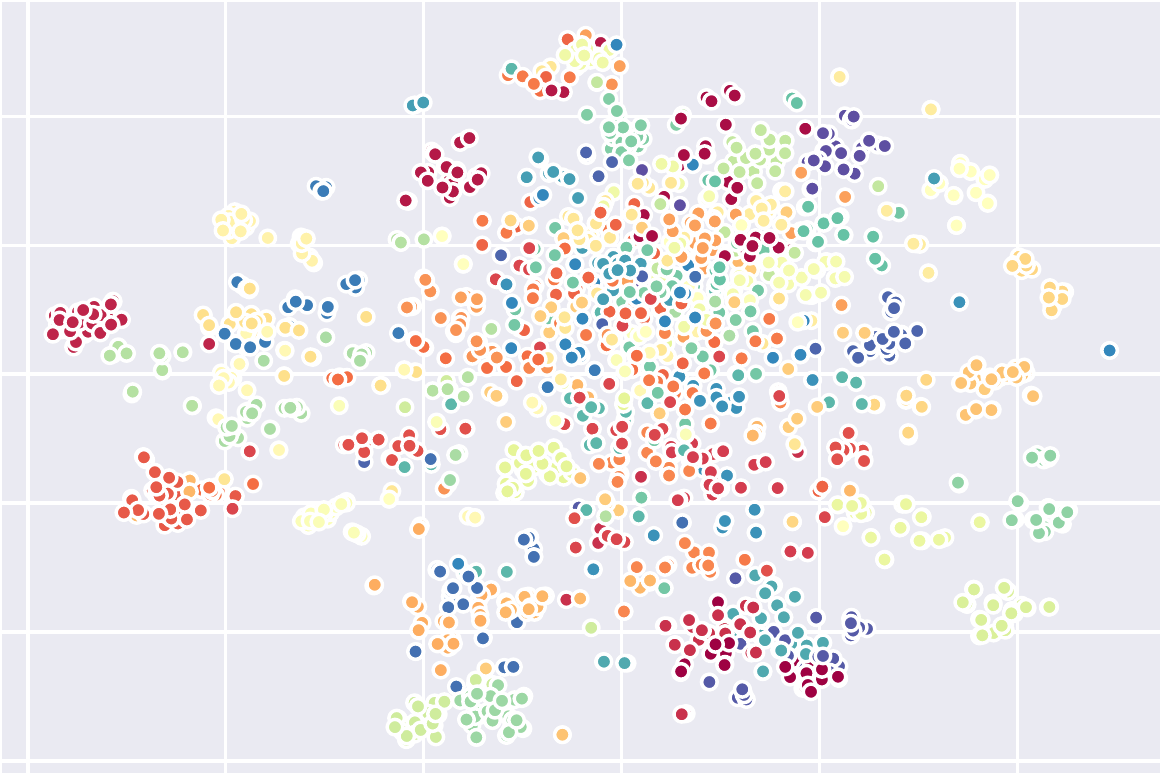}
  \caption{}
  \label{fig:tsnesfig4}
\end{subfigure}
\caption{t-SNE embeddings of HMDB test videos from networks trained on various data. Each color represents a different activity. (a) Randomly initialized network (b) ImageNet trained network (c) Kinetics trained network (d) Our evolved loss.}
\label{fig:fig}
\end{figure*}

\begin{figure*}
\begin{subfigure}{.5\textwidth}
  \centering
    \includegraphics[width=0.9\linewidth]{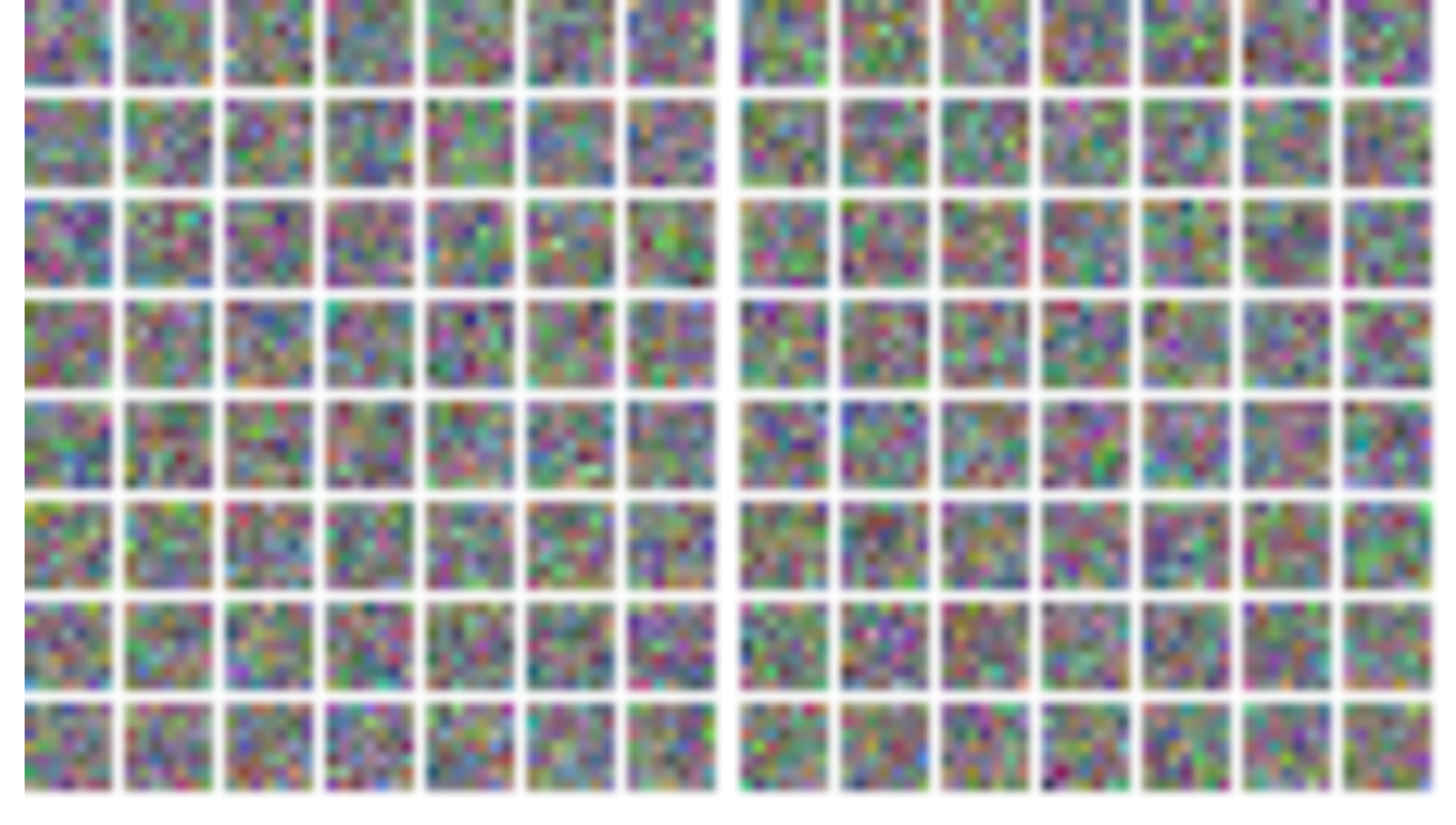}
  \caption{}
  \label{fig:filtersfig1}
\end{subfigure}%
\begin{subfigure}{.5\textwidth}
  \centering
    \includegraphics[width=0.9\linewidth]{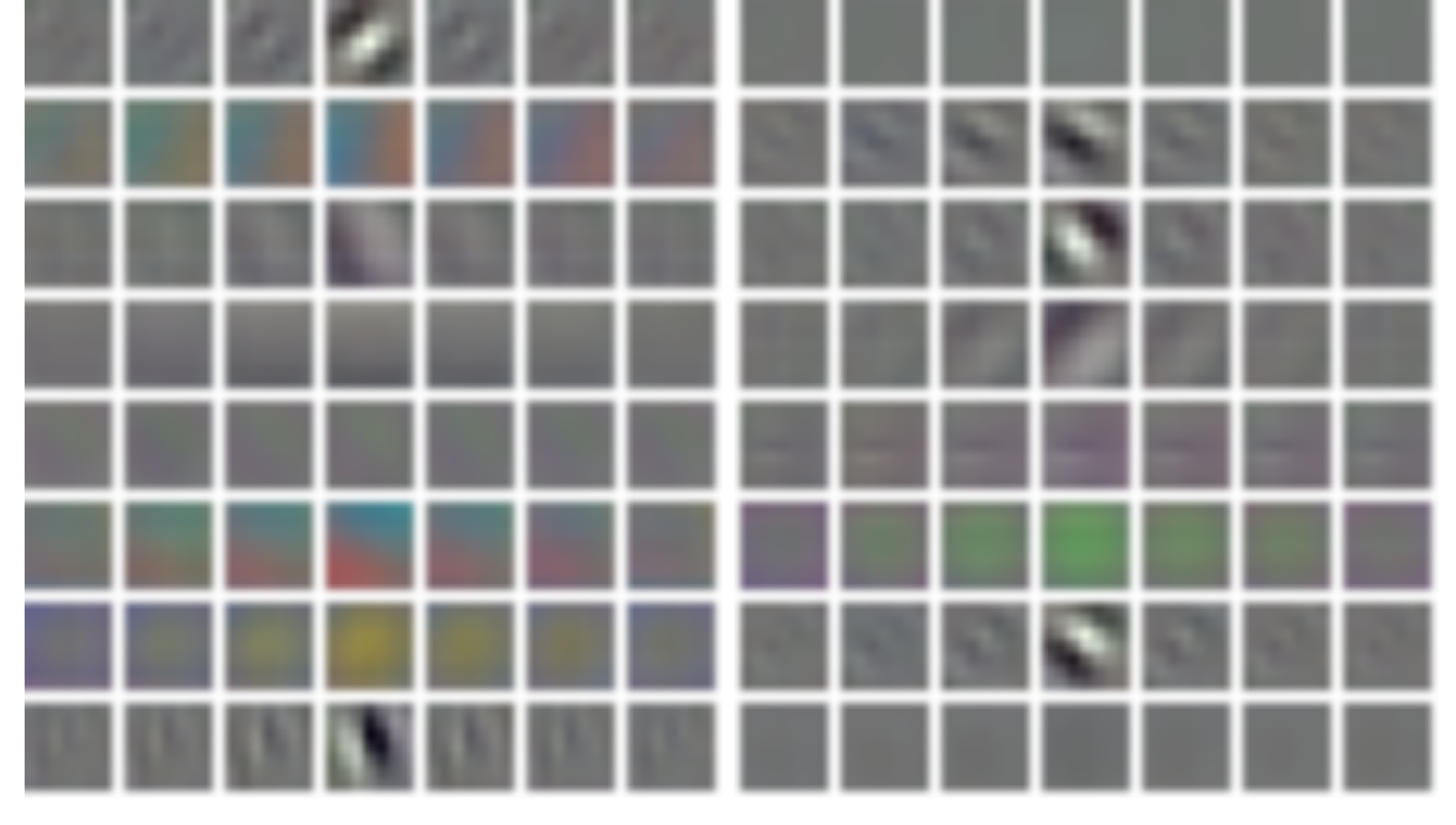}
  \caption{}
  \label{fig:filtersfig2}
\end{subfigure}
\begin{subfigure}{.5\textwidth}
  \centering
    \includegraphics[width=0.9\linewidth]{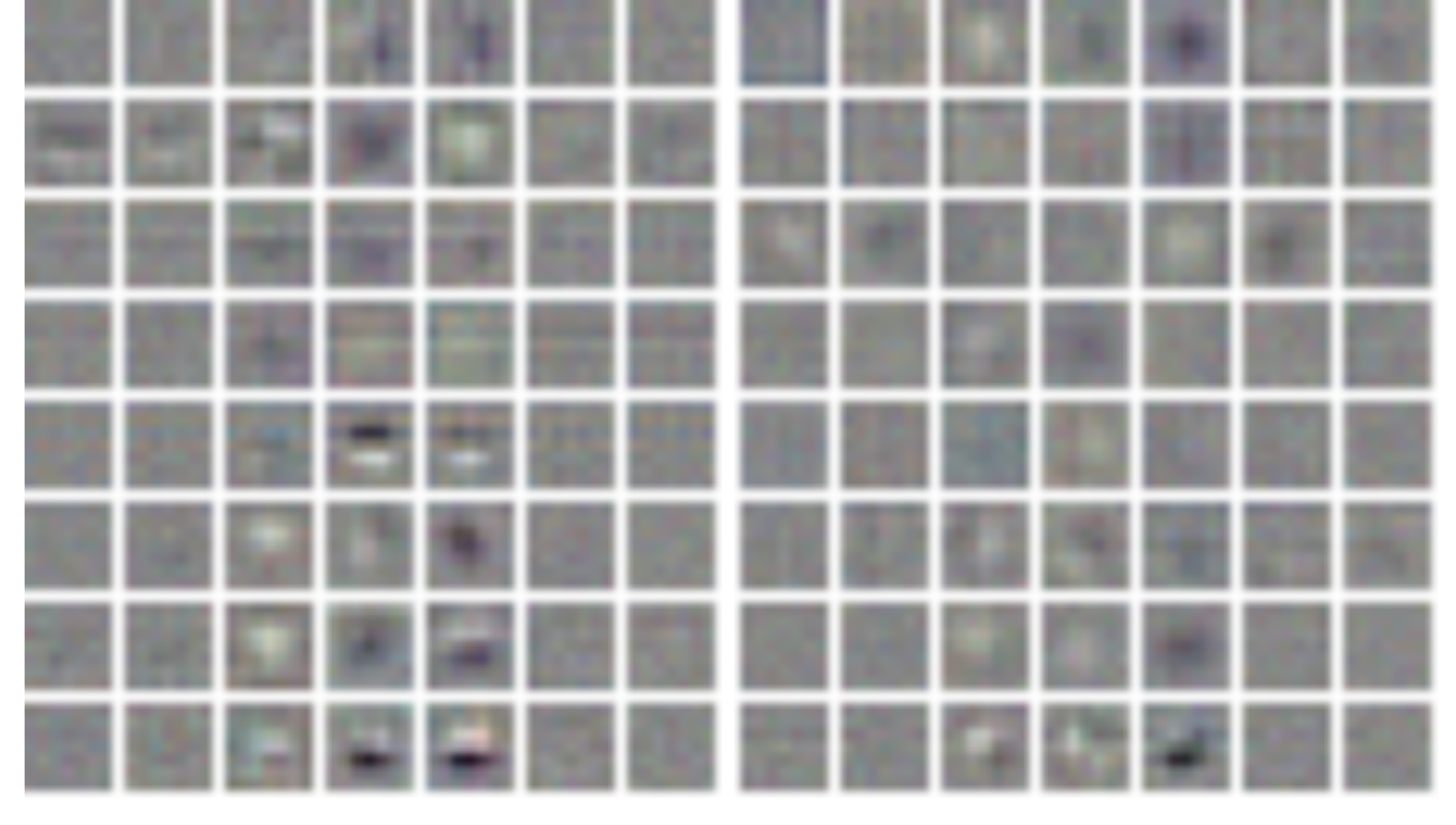}
  \caption{}
  \label{fig:filtersfig3}
\end{subfigure}%
\begin{subfigure}{.5\textwidth}
  \centering
    \includegraphics[width=0.9\linewidth]{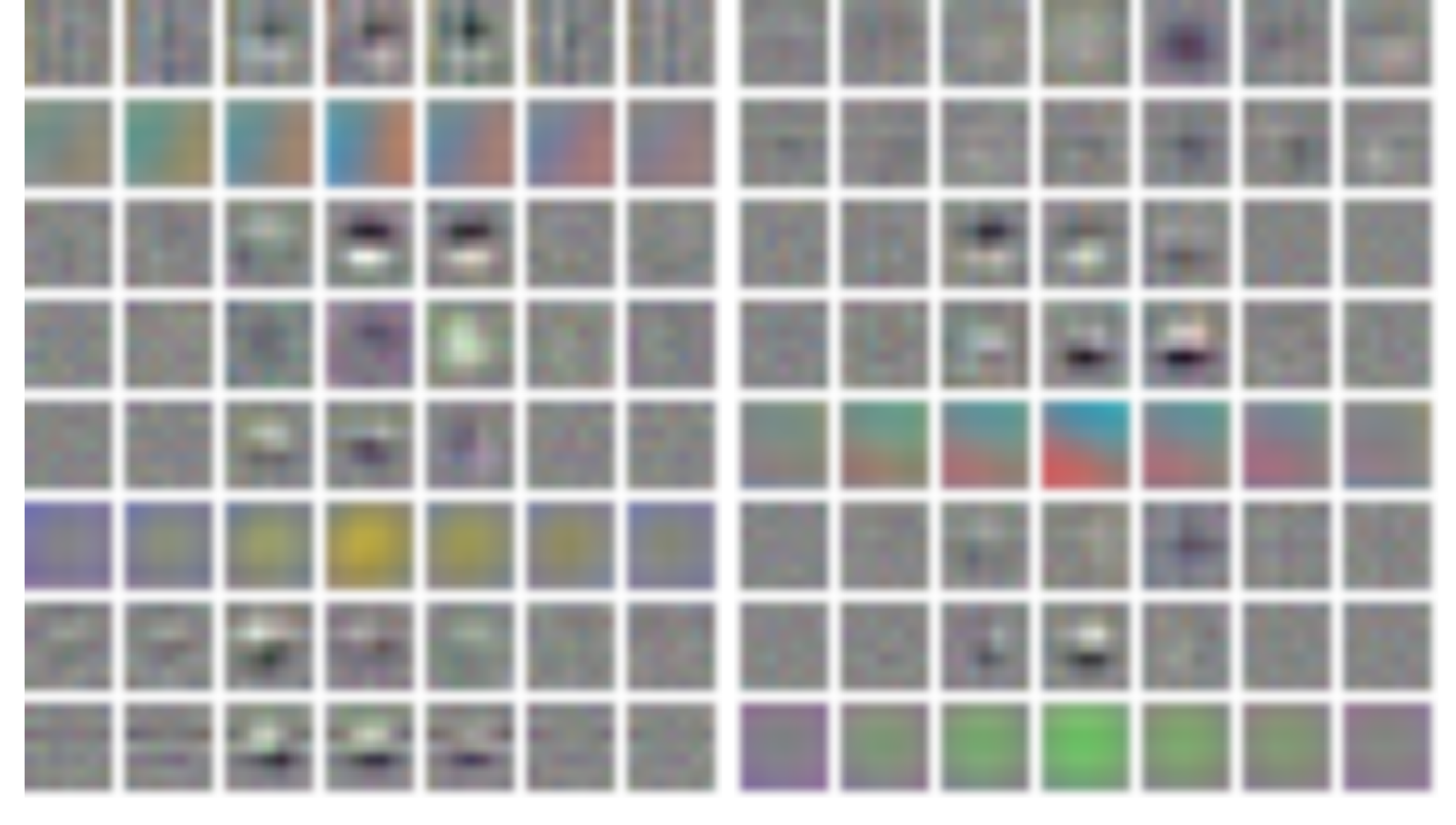}
  \caption{}
  \label{fig:filtersfig4}
\end{subfigure}
  \caption{Visualization of (a) random filters (b) filters learned with standard supervised learning and (c) AVTS \cite{korbar2018cooperative} self-supervised learned filters (d) filters learned with our evolved multi-modal, multi-task loss function.}
  \label{fig:filter-vis}
\end{figure*}

\section{Datasets}
We compare our approach, referred to as ELo in the paper, on 3 standard video recognition datasets. Kinetics \cite{carreira2017quo} is a large-scale video dataset with over 200k labeled video clips for 400 different activity classes. Each clip is 10 seconds long, leading to over over 500 hours of annotated video data. HMDB \cite{kuehne2011hmdb} is a smaller dataset with around 3000 training and 1500 validation video clips for 51 different activities. On average, each video is 3 seconds long. UCF-101 \cite{UCF101} is similar to HMDB with 101 different actions, and about 13,000 videos split into training and test sets.

Using both large-scale data and smaller datasets shows that the representation obtained from unsupervised learning is general and works well even with limited labeled data.

\section{Visualization of loss evolution}
Fig. \ref{fig:fig} shows the t-SNE embedding for our unsupervised approach compared to random weights, ImageNet trained CNNs, and Kinetics trained CNNs. ELo  generates more clear video clusters than random weights and ImageNet weights, and is more comparable to the model trained with supervised Kinetics videos and labels.


\section{Supplemental Results}


Fig \ref{fig:filter-vis} visualizes the filters our approach learned compared to other approaches: it shows filters from random initialization, supervised learning with large data, previous self-supervised learning, and our unsupervised learning method ELo. ELo filters are quite similar to those learned with labeled data, but do show some differences.

\end{document}